\documentclass[conference,UTF8]{IEEEtran}

\IEEEoverridecommandlockouts
\usepackage{cite}
\usepackage{amsmath,amssymb,amsfonts}
\usepackage{algorithm}
\usepackage{algcompatible}
\usepackage{algpseudocode}
\usepackage{graphicx}
\usepackage{textcomp}
\usepackage{xcolor}
\usepackage{CJKutf8}
\usepackage{subcaption}
\usepackage{caption}
\usepackage{hyperref}
\usepackage[normalem]{ulem}
\usepackage{flushend}

\def\BibTeX{{\rm B\kern-.05em{\sc i\kern-.025em b}\kern-.08em
    T\kern-.1667em\lower.7ex\hbox{E}\kern-.125emX}}

\newcommand{\edit}[1]{#1}

\title{ Efficient Exploration via First-Person Behavior Cloning Assisted Rapidly-Exploring Random Trees}

\author{\IEEEauthorblockN{Max Zuo\IEEEauthorrefmark{1} \begin{CJK*}{UTF8}{gbsn}(左绍文)\end{CJK*}}
\IEEEauthorblockA{zuo@gatech.edu}
\and
\IEEEauthorblockN{Logan Schick\IEEEauthorrefmark{1}}
\IEEEauthorblockA{lschick3@gatech.edu}
\and
\IEEEauthorblockN{Matthew Gombolay\IEEEauthorrefmark{1}}\thanks{\textit{* School of Interactive Computing}, \textit{Georgia Institute of Technology}}
\IEEEauthorblockA{matthew.gombolay@cc.gatech.edu}
\and
\IEEEauthorblockN{Nakul Gopalan\IEEEauthorrefmark{1}}
\IEEEauthorblockA{ngopalan3@gatech.edu}}

\begin{document}

\maketitle

% \begin{abstract}
% \ngnote{Paper in incorrect format. HRI is using IEEE (Check the requirement) and double blind (remove author names)}.
% As game environments get bigger, so do the requirements for exploration testing. 
% \ngnote{Is there previous work in this area? Why is it not sufficient?}
% In this paper we analyze the effectiveness of Rapidly-exploring Random Tree (RRT) algorithms in regards to game exploration. This paper  \ngnote{introduces a novel algos... and} analyzes the effectiveness of \underline{h}uman-\underline{a}ssisted RRT (H\edit{S}-RRT) and behavior-\underline{c}loning \underline{a}gent RRT (CA-RRT) algorithms with respect to the number of game states searched and the time taken to explore those game states. These methods can lead to significant time and compute savings compared to a naive method in exploring games. Using custom-made gym-minigrid environments, we find H\edit{S}-RRT (and in many cases, CA-RRT) to explore more game states in less searches when compared to the vanilla RRT algorithm.
% \ngnote{Numerically how well do we do? No surprises to the reader.}
% \end{abstract}

% TODO: should mention human guided sparse reward function via demonstration

\begin{abstract}
Modern day computer games have extremely large state and action spaces.
To detect bugs in these games' models, human testers play the games repeatedly to explore the game and find errors in the games. 
Such game play is exhaustive and time consuming. 
Moreover, since robotics simulators depend on similar methods of model specification and debugging, the problem of finding errors in the model is of interest for the robotics community to ensure robot behaviors and interactions are consistent in simulators.  
% This work attempts to help game testing by automating game exploration using ideas fr
% To reduce the burden on human testers previous methods have considered search algorithms~\cite{zhan2018taking, 2018arXiv181106962D} or reinforcement learning (RL) based exploration methods~\cite{gordillo2021improving} to find bugs by exploring a game's state space automatically. 
Previous methods have used reinforcement learning~\cite{gordillo2021improving} and search based methods {\cite{chang2019reveal, Chang21Boosting}}~\cite{2018arXiv181106962D} including  Rapidly-exploring Random Trees (RRT) to explore a game's state-action space to find bugs.
% Previous methods have used Rapidly-exploring Random Trees (RRT) to explore a game's state action space~\cite{zhan2018taking} as the RRT search tree is biased to explore unexplored states~\cite{rrt}. 
% RRTs are ideal for searching large continuous state-action spaces as during each iteration of the algorithm the search tree is biased to explore unexplored regions of the state space~\cite{rrt}.
However, such search and exploration based methods are not efficient at exploring the state-action space without a pre-defined heuristic.
In this work we attempt to combine a human-tester's expertise in solving games, and the exhaustiveness of RRT to search a game's state space efficiently with high coverage. 
% As game environments get bigger, so do the requirements for exploration testing. 
% \ngnote{Is there previous work in this area? Why is it not sufficient?}
% \edit{Prior} works use traditional search methods~\cite{zhan2018taking}~\cite{2018arXiv181106962D} or reinforcement learning (RL) based exploration methods~\cite{gordillo2021improving} \edit{for game exploration to help detect bugs}. Traditional search can be exhaustive in both time and memory, and may require expert-designed heuristics or weights \edit{to efficiently explore a game state space,} which may not always be available. \edit{I}ntrinsic motivation may help alleviate situations with sparse reward~\cite{burda2018exploration}, current methods in game testing with RL still require a significant amount of hand-holding~\cite{gordillo2021improving}. 
% In this paper, we analyze the effectiveness of Rapidly-exploring Random Tree (RRT) methods in regards to game exploration. \edit{RRT methods, while primarily used for pathfinding, can be repurposed for wide exploration by removing the goal configuration in search and simply focusing on expanding~\cite{zhan2018taking}.} 
{This paper introduces \underline{C}loning-\underline{A}ssisted RRT (CA-RRT) to test a game by searching through game states. 
% and the compare different methods by the time taken to explore those game states. 
We compare our methods to two existing baselines: 1) a weighted-RRT as described by Zhan et. al.~\cite{zhan2018taking};  2) human-demonstration seeded RRT as described by Chang et. al.~\cite{chang2019reveal}.
% i.e., creating search nodes in a planner, in our case RRT, for human demonstrated states to improve game exploration as done by . 
% These methods can lead to significant time and compute savings compared to a naive method in exploring games. 
% On custom-made\edit{,} gym-minigrid environments, 
We find that CA-RRT is applicable to more game maps and explores more game states in \edit{fewer} \edit{tree expansions/iterations} when compared to the existing baselines. In each test, CA-RRT reached more states on average in the same number of \edit{iterations} as weighted-RRT. In our tested environments, CA-RRT was able to reach the same number of states as weighted-RRT by more than $5000$ fewer \edit{iterations} on average, almost a $50\%$ reduction and applied to more scenarios than. Moreover, as a consequence of our first person behavior cloning approach, CA-RRT worked on unseen game maps than just seeding the RRT with human-demonstrated states.} 
% Moreover, CA-RRT is able to search through novel maps when compared to the seeded RRT as it does not have any
% \ngnote{Numerically how well do we do? No surprises to the reader.}
\end{abstract}

\section{Introduction}

% motivation/what we are trying to solve

There is an explosion in model specifications for games, and simulators, be it for entertainment, or to save on physical learning interactions for robots. However, testing for these models is still performed manually which becomes challenging with large, continuous, state-action spaces. 
The problem of such model testing is well studied in computer games, which will be the focus of this paper. However, our algorithms can generalize to robotics simulators as they can learn and explore in continuous state-actions spaces.
% Game testing is a pivotal part of game development. 
Testing a game's  model or ``play'' involves extensive manual testing, where human players play the game multiple times and try to reach every possible state with the purpose of reporting bugs and inconsistencies. 
However, \edit{as games continue to grow in complexity, with larger action and state spaces,} it is no longer feasible to test a game's model using human game play alone. 
In this paper we propose a method that learns heuristics for a search based algorithm using a few human demonstrations. 
These heuristics allow the search algorithms to automatically test other states in the game removing the necessity to test all possible scenarios in the game using human game plays.  
% Therefore, research has been conducted on using various forms of automation techniques to complement the manual game testing process.
% It is technically possible to solve this problem of game exploration through brute force using random actions. As we will see, there is literature that does just this. \ngnote{cut the previous sentence, just add the citations to the line before that.}

% General mention of previous works
Previously, search based methods have been used to find goals in a task quickly using their capacity to explore~\cite{griner2012human}~\cite{caves2010human}~\cite{HG-RRT}~\cite{HA-RRT}.
However, traditional search based algorithms can take large amounts of time and compute power when a specific sequence of actions is required to advance. 
For example, in most games the action to \emph{open a door} is allowed in all states. However, a door would only open as a result of this action when the agent has taken actions to pick up a key already, and be present next to a door. Search algorithms spend a lot of compute and time to discover these long sequence of actions. 
% This is especially true in states where only a specific sequence of actions will allow the agent to explore more states.  
% For example, search methods may exhaustively explore a large action space when only a few actions per state \edit{can modify the state itself}. 
A human tester \edit{can select} these actions rather trivially in comparison.
Previous methods have attempted to use human capacity to advance in games using expert-programmed heuristic~\cite{2018arXiv181106962D} or by using a pre-specified distribution of human preferences~\cite{zhan2018taking}. {Some works have seeded an RRT search tree using states seen in the human demonstration\cite{chang2019reveal}, and others have used human demonstrations with behavior cloning to learn the human's intent~\cite{Chang21Boosting}. However, these methods fail to generalize to novel unseen maps as these policies depend on an agent's underlying states of location and orientation in the game. Reliance on the underlying state of an agent makes these approaches brittle as this underlying state is not always available and its distribution can change between game maps.} 
% as they depend on knowing an agent's underlying state in the map of a game, which reduces their applicability in novel scenarios.}  
% but automated methods require a predefined standard distribution \ngnote{standard what? We are not talking about some Gaussians here....}~\cite{zhan2018taking} or an expert-programmed heuristic~\cite{2018arXiv181106962D}. 

% In this paper we develop \edit{HS-RRT, adapted from a family of human-RRT collaboration algorithms}~\cite{griner2012human}~\cite{caves2010human}~\cite{HG-RRT}~\cite{HA-RRT} for gameplay-testing. 
{In this paper we develop Clone Assisted RRT (CA-RRT) which uses first-person demonstrations with behavior cloning to generalize to novel maps from where RRT can be used to explore a game allowing generalization to unseen states.
% We then also propose the behavior-cloning assisted RRT algorithm\ngnote{CA-RRT?}, both of which utilize human player demonstrations. 
We compare our algorithm to previous baselines such as weighted-RRT~\cite{zhan2018taking} and human-seeded RRT~\cite{chang2019reveal} on their ability to find novel states in the least number of RRT node expansions. 
% These human-guided RRT search techniques can also help robots find goal conditions more efficiently~\cite{griner2012human}~\cite{caves2010human}, but the focus of our work here is to show the gains of RRT with human demonstrations in exploring state spaces with fewer iterations.
% When comparing the completeness of game exploration methods, it is important to have a ground truth for how many states have been visited, which is not always immediately available. 
% We therefore also analyze the exploration metrics used in previous work~\cite{zhan2018taking} and propose a metric that is closer to the ground truth state exploration. 
In our experiments, we find that CA-RRT requires about half as many node expansions to search the entire search space when compared to the weighted RRT method~\cite{zhan2018taking} and applies to more maps than just seeding an RRT tree with human-demonstrated states~\cite{chang2019reveal}. }
% outperform other algorithms, sometimes cutting the number of expansions by half. \edit{In the new \textbf{DualHallway} gym-minigrid~\cite{gym_minigrid} environment, on average CA-RRT saturated the game state space in 6836 iterations, which is 5831 iterations fewer than weighted RRT, and in the \textbf{CascadingLockDoor} environment CA-RRT saturated the game state space in 6659 iterations, which is 5933 fewer iterations than our baseline weighted RRT.}
Our contributions in this work are:
\begin{enumerate}
    \item {CA-RRT: which uses first person behavior-cloning and RRT to explore game states randomly given novel unseen states.} 
    \item Results showing CA-RRT outperforms previous weighted RRT baseline~\cite{zhan2018taking} by using fewer node expansions to explore the entire state space in two gym-minigrid domains:  \textbf{DualHallway} $6836$ expansions vs  $12667$, and \textbf{CascadingLockDoor} $6659$ vs $12592$.
    \item {Results showing CA-RRT applies to more maps than RRT methods dependent on knowing the underlying global location and orientation state of the agent.}
\end{enumerate}

% what kinds of experiments (environment, numerical results)

% \ngnote{Write this intro in a four part structure. 1) What are we trying to solve. 2) What has been done previously. 3) What is missing? What are we doing to fix it. 4) A snapshot of our results, numeric, so readers get an idea of what to expect. It is also advisable to end the intro section with a list of contributions. It makes reading the paper easy.}

\section{Related Work}
% \ngnote{Related work section too large. It needs to be between 0.5 - 1 page. No more. Do not use subsections here. It uses up a lot of space. }
Games and simulators are becoming pervasive in entertainment and automation. These systems require testing to verify their model specification. Automatic testing tools are becoming common to help in game testing and verification.
% , which has led to a diverse set of techniques being applied to the problem.
% Recent literature has found RRT to be an algorithm with a lot of potential for the game testing problem~\cite{zhan2018taking}.
% \subsection{Video Game Exploration}
% \ngnote{Think of each citation getting about a line each. Remove the subsections use bold with paras if you need. Also store citations with last-name\_year in the bibtex according to the first authors. This way you can type by memory. May be name one or two games , but this is too much detail here.}
% The video game market is a billion dollar industry~\cite{9234724}. As the scale of games and their revenue grow, so do users' expectations in terms of verification and quality assurance. Due to the importance of game testing and its need in the gaming industry, there has been increasing literature focusing on the automatic testing of games.
Aghyad and Moataz~\cite{9234724} explore a wide breadth of these methods, including evolutionary algorithms, genetic algorithms, and graph search algorithms for searching game state spaces.
% Ubisoft uses artificial intelligence agents to balance, test, and enhance the levels in their games. 
The Ubisoft development teams for ``Tom Clancy's: The Division" implemented bots to test the game servers and to test procedurally-generated levels in the game respectively~\cite{web:lang:stats}. Electronic Arts uses A* \edit{bots to automate goal-driven gameplay} to find discrepancies and flaws within their games~\cite{2018arXiv181106962D}. Curiosity-driven reinforcement learning~\cite{pathak2017curiosity} agents have also been used to improve gameplay coverage for testing~\cite{gordillo2021improving}.
Search algorithms like A* are good at finding paths to an end-goal given a pre-specified heuristic. Such end-goals and heuristics are hard to specify for large games. 
% A*, while proficient in pathfinding and navigation, also requires domain knowledge in defining a proper, functional heuristic function\edit{, and is not particularly well-suited for wide exploration}~\cite{2018arXiv181106962D}. %While all these automatic testing strategies have been used in modern game development, one search based approach has been shown to outshine the rest. This approach has to deal with rapidly-exploring random tree search.
% \subsection{RRT}

Rapidly-exploring random trees (RRT) is traditionally used for path planning~\cite{rrt}. 
An RRT is biased to explore large unexplored regions in the state-space to find a path to the goal location~\cite{rrt}.
% However, its property of exploring search spaces can also be used for covering large state spaces and are especially applicable in continuous spaces. 
\edit{By removing the goal configuration in search and simply focusing on expanding, RRT can be utilized to explore large state spaces \emph{without an end goal}}. Such exploration helps cover all possible states in the game helping test corner cases that can be reached by a player.  
% Alternatively, RRT for exploration can be thought of as }
% Although RRT does not seem to have been used in industry yet, some researchers are finding it to be valuable in game exploration.
RRTs have been used to explore Super Mario World~\cite{zhan2018taking} among other games ~\cite{inproceedings}. 
Instead of trying to optimize game score or directly optimizing game state coverage, Zhan et al.~\cite{zhan2018taking} focused on maximizing diversity of game states visited in an analogous way to how some RL intrinsic motivation methods work~\cite{burda2018exploration}~\cite{pathak2017curiosity}.

Some researchers have looked into using human input as guidance for search or planning algorithms~\cite{caves2010human}~\cite{griner2012human}~\cite{HA-RRT}~\cite{HG-RRT}: HA-RRT and Human-Guided RRT (HG-RRT*), for example, utilize human expert specifications for assisting exploration of a virtual urban environment and motion planning~\cite{HA-RRT}~\cite{HG-RRT}. Zhan et al.~\cite{zhan2018taking} find that human play-testers explore rapidly initially but exploration quickly fatigues, and over sufficient amount of time, RRT can overtake humans in number of states explored. In this work we aim to combine both these methods via learning heuristics for  search algorithms.  
{Chang et al. \cite{chang2019reveal} present a method which uses  demonstrated human trajectory states as configuration states in RRT. This allows fast exploration from previously visited states. Chang et al.~\cite{Chang21Boosting} further extend this approach to a behavior cloning based approach with the agent's global position to explore the state space from generalized policies rather than just configuration states. We improve upon this approach by using a first person image based agent state to learn policies that generalize to novel maps.} 
% We hypothesize that reducing the effective branching factor of RRT will allow \edit{RRT} to more effectively \edit{explore} the game state space, whether through human demonstrations or through a behavior clone learning from human actions. 
Even though a behavior cloning may not be able to complete an environment by itself, we believe the clone will be able to efficiently support the search process, similar to DeRRT*~\cite{kuo2018deep} and AlphaGo~\cite{alphago}. These previous methods used demonstrations to reduce the effective branching factor for search allowing an agent to find a goal fast. We, on the other hand, will use the demonstrations to seed novel RRT trees to search the state space rapidly for bugs, and we are not looking for an individual goal state but to ensure that there are no bugs in any of the states of a game or simulator.  
% used a similar mechanism to reduce the effective branching factor for Monte Carlo tree search \edit{in the board game Go}. However, instead of maximizing a reward function for winning, we wish to improve gameplay testing through efficient wide exploration.

\section{Methodology}
In this section we describe the CA-RRT algorithm. 
Previous algorithms that use RRT for exploration such as HA-RRT~\cite{HA-RRT} and HG-RRT*~\cite{HG-RRT} allow guidance from anywhere in the configuration space. 
They do so by allowing users to specify waypoints, avoidance points and drawing guidance paths. However, these methods do not allow direct modification or influence of the RRT trees and therefore cannot fully take advantage of human demonstrations. Furthermore, HG-RRT* waypoints or HA-RRT paths must be redefined or redrawn for every small change in the environment, leaving humans tethered to the search process. 
{Chang et al.~\cite{chang2019reveal} use states from human demonstrations as 
seed configurations states within an RRT. We refer to this idea as human-seeded RRT (HS-RRT) and improvise on it by adding a first person behavior cloning based approach over it.}
\subsection{CA-RRT}
\label{algorithm1}
\begin{algorithm}\caption{CA-RRT Exploration}

    \begin{algorithmic}[1]
        \Function{CA-RRT-EXP}{$env, initial\_config$, $\pi_{\textrm{BC}}$}
        % \State $Goals \gets env.states$
        \State $visited \gets \{initial\_config\}$
        \For{$K$ times}
        \State Rollout $\pi_{\textrm{BC}}$ for $N$ steps into a buffer
        \EndFor
        \While {$\neg \texttt{done}$}\Comment{Collect buffer trajectory and add to visited configurations}
        \State $action \gets buffer.read\_action()$
        \State $env.step(action)$
        \State $current\_configuration \gets env.conf()$
        \State $visited \gets visited \cup \{env\}$
        \EndWhile
        \State $\texttt{RRT}(env, visited)$
        \EndFunction
    \end{algorithmic}
\end{algorithm}

\edit{H\edit{S}-RRT requires collecting human trajectories and using human-visited states as seed nodes in its tree. Moreover, H\edit{S}-RRT can only be run on an instantiation of a game where a human has played on,} limiting our ability to test different initial conditions or even slightly changed game configurations\edit{: we would need to collect a human trajectory in each version of an edited game}.
{The limiting factor here is the use of global state information from the human trajectory that directly seed the configuration space of the RRT algorithm. This information does not generalize to novel maps with different global state information or even a change in the origin co-ordinate location within the same map. Extensions that use global information to train a behavior cloning policy such as Chang et al.~\cite{Chang21Boosting} suffer from the same issues.
To ensure that our agent can explore unseen environments we developed CA-RRT, which is a first-person image-state-based (behavioral) Clone-Assisted RRT searching algorithm. }

CA-RRT starts with collecting a set of human-created trajectories. {We train a behavioral cloning policy using the first person state-action pairs from those trajectories}. Before RRT runs on the environment, we use the behavior cloned policy to rollout for some number of timesteps similar to H\edit{S}-RRT \edit{and use the clone-collected trajectory of state-action pairs} to seed our search {as seen on line $4$ of~\hyperref[algorithm1]{Algorithm 1}}. \edit{Because CA-RRT produces its own trajectories, there is no reliance on a human operator to seed the search each time as is the case with H\edit{S}-RRT. CA-RRT can therefore run with different initial configurations and on environments where the clone was not trained on at all, as it does not need a human player to collect trajectories after the initial training phase}. During RRT, the clone agent produces the action probabilities for every state which RRT uses as a prior to sample from.
% \ngnote{Cite this earlier in the related work section. There is also a work that modifies RRT exploration w.r.t. language commands and the current state: ``Deep sequential models for sampling-based planning'' - Yen-Ling Kuo, Andrei Barbu, and Boris Katz.  We want to say that this work is improving Game state Testing by wide exploration, and not looking to maximize an objective function like Alpha Zero and other methods.}.

 %\ngnote{more details about the size of the layers after each layer. How many is a couple? Be precise.}

During evaluation of CA-RRT, we still wish to fully explore the range of game states, not simply the expected trajectory learned by the behavioral cloning training. Therefore, we use a Laplace smoothing over the action probability distribution, and the factor $\alpha$ increases after every step allowing the agent to choose actions more uniformly.

\section{Experiments}
Our work is evaluated in the OpenAI Gym~\cite{gym} library with custom gym-minigrid~\cite{gym_minigrid} environments built  to be extremely challenging if not given a detailed and extrinsic reward function. 
{We compare the results of our \edit{C}A-RRT method to the vanilla RRT method from Zhang et al.~\cite{zhan2018taking} and \edit{the H\edit{S}-RRT method} from Chang et al.~\cite{chang2019reveal}}. 
% The new environments are publicly available in a forked repository. 

\subsection{Gym-minigrid}

Gameplay-testing aims to evaluate and test as many reachable game states as possible, if not all of them. In many environments, the number of game states varies from one play to another. Often times the number of game states can be immeasurably large: the game state could be a continuous space, or the branching factor given the number of possible actions explodes exponentially.
In order to conduct controlled experiments with calculable ground truth state coverage, we built a set of our own environments on top of the gym-minigrid~\cite{gym_minigrid} library with a countable number of states. {Gym-minigrid has previously been been used in other game-testing work such as Chang et al.~\cite{Chang21Boosting}. We create novel gym-minigrid environments in this work to test the different approaches.}
% The components of our environments are minimal such that the individual map locations the agent visits sufficiently represent information about game states. This allows us to evaluate game state coverage by measuring the game map coverage. However, the RRT methods explored here are not limited to just game map coverage. \edit{For example, Zhan et al., used neural image embeddings to explore game state coverage instead\cite{zhan2018taking}}.
Gym-minigrid~\cite{gym_minigrid} describes a family of lightweight environments built for tasks such as reinforcement learning. Every environment has the same basic building blocks: the agent, doors - which may require a key to open, walls, balls, etc. The agent can only pick up one \edit{item (e.g. key, ball)} at a time. The agent is provided with a partial observation of the state of the game from its perspective, much like how a first-person game limits what the human player can view. We also created a new, tougher building block which we employ and recommend using, the \textbf{MultiLockDoor}, which requires several keys to open. %, not just one~\cite{max_gym_minigrid}. The color of the locked door is associated with one of the keys required to unlock it, but the other colors are not specified~\cite{max_gym_minigrid}. The other key colors can be specified in the mission string or may remain unspecified. We hope to add the ability for the door to change colors to a new lock color once the corresponding key with the same color is applied to the door. However, this adds a layer of difficulty which we hope will further illustrate the difference between what H\edit{S}-RRT can accomplish in gameplay-testing tasks over current methods.
\begin{center}
    \label{fig1}
    \begin{minipage}{0.15\textwidth}
        \centering
        \includegraphics[height=0.3\textheight]{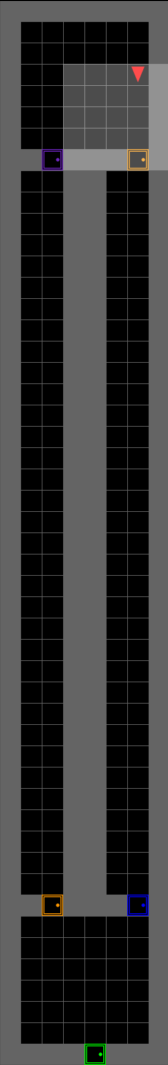}
    \end{minipage}
    \begin{minipage}{0.15\textwidth}
        \centering
        \includegraphics[height=0.12\textheight]{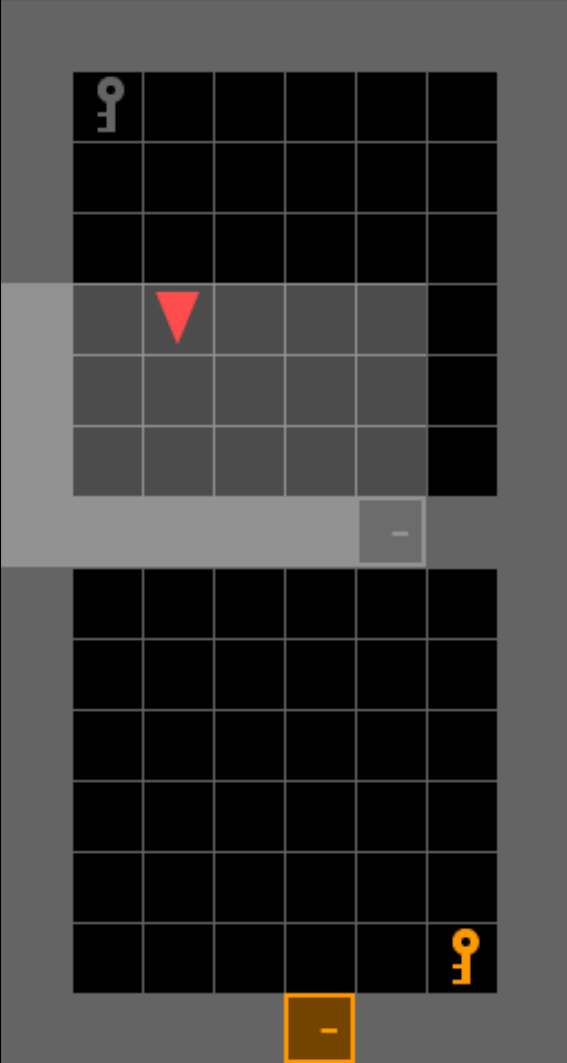}
    \end{minipage}
\end{center}
\begin{center}
    \captionof{figure}{The environments tested on. From the left: \textbf{DualHallway-v0}, \textbf{CascadingLockDoor-v0}.}
\end{center}

\subsubsection{DualHallway Environment}
The first environment experimented in is the \textbf{DualHallway} environment which invites a playing agent to open a few doors to reach the end goal room through paths that can include two possible locked hallways. In order to cover a large portion of the game state space effectively, a game play testing agent must efficiently explore all hallways and rooms, which will require the agent to use the \textit{toggle} action. These doors are not locked. Door location, door color, agent orientation, and agent position are all randomly assigned. \edit{See example in \hyperref[fig1]{Figure 1}.}

\subsubsection{CascadingLockDoor Environment}
This environment incorporates the \textbf{MultiLockDoor} object. There two rooms, where the gameplaying agent is asked to reach the end of the second room. In each room, there is a key which corresponds to the door at the end of the room, which the gameplaying agent must pick up and use on its corresponding door. The first key must be used on the second lock as well. \edit{See example in \hyperref[fig1]{Figure 1}.}

% \subsubsection{LockedHallway Environment}
% The last environment tested on. This environment also has the gameplaying agent try to reach the end of a sequence of rooms. However, in this environment, all the keys are placed in the first room, and each key is associated with the color of each door. The agent must select the correct key and apply it to the door for it unlock~\cite{max_gym_minigrid}. Only one key is needed per door.

\subsection{Experiment Setup}
For both of these environments, a single trajectory by a human player was collected and used to train a crude behavioral cloning model. We then run RRT, H\edit{S}-RRT, and CA-RRT on each of these environments and measure how quickly \edit{each algorithm} was able to saturate, i.e., explore all states of, the game state space. \edit{RRT*~\cite{karaman2011sampling} is not explored in here as the main goal is not to find the optimal path through a configuration space, but to rather explore through the entire space.} 
% In order to measure game state space coverage, we opt to use map coverage, rather than nuclear norm of neural network embeddings~\cite{zhan2018taking}. Since these environments exist in a discrete gridworld and a discrete action space with limited interactions, the game map space tells us almost everything about the game state for our environments. 
We can now directly calculate the upper limit of reachable states \edit{by counting the number of discrete possible reachable game map locations}. \edit{Game map coverage} can still be used in a continuous state space, if we were to compute a game state hash to represent whether or not a state\edit{/discretized map location} has been reached before.

The CA-RRT policy is a convolutional neural network with an input of images with the size $(56,56,3)$. The architecture used is a convolutional layer, followed by a residual block, repeated, followed by two fully-connected layers both with $16$ output nodes, and a final layer with an output size of $7$, the number of actions allowed. A dropout layer is used with a dropout rate of $10\%$. Any similar learner would have been mostly equivalent in effectiveness.
For each experiment, the agent position and orientation, door color and position, and key position are all permuted. Therefore, CA-RRT and RRT were never evaluated on the exact instantiation of an environment seen during training time. For \textbf{DualHallway} environments, the Laplace smoothing factor was $\alpha=0.1$ which increased at a rate of $10^{-5}$ every iteration, and for \textbf{CascadkingLockDoor}, $\alpha=0.1$ with a growth rate of $10^{-3}$. Once $\alpha\ge0.5$, CA-RRT defaulted to hand-defined weights. The hand-crafted weights used may not be optimal. However, \edit{the weights} were designed by domain knowledge, and empirically chosen for their performance.

% We also explicitly incorporate a difference between assuming a non-deterministic environment and a deterministic environment. In a non-deterministic environment, we assume that any action taken may result in a random successor state, and that taking the same action multiple times from the same initial state may result in a different successor state each time. This is the assumption the authors using RRT and the nuclear norm metric make in~\cite{zhan2018taking}.

We differentiate between versions of RRT where actions are sampled uniformly versus RRT where actions are sampled over a user-defined distribution. Building a user-defined distribution requires domain knowledge of the problem, and is explored in prior works~\cite{zhan2018taking}. 
% For generality \edit{with respect to prior works}, we assume stochasticity despite \edit{knowing our environments are deterministic here because possible instances of our domains can be stochastic}~\cite{zhan2018taking}. 
% Experiments show assuming a deterministic environment \edit{(in deterministic environments, which is not always the case)} can drastically speed up each of the tested algorithms, \edit{however, we test assuming stochasticity because many gaming environments are indeed stochastic}.
% \subsubsection{Distractions}
\begin{center} \label{figure_ddb_example}
    \includegraphics[width=0.15\textwidth]{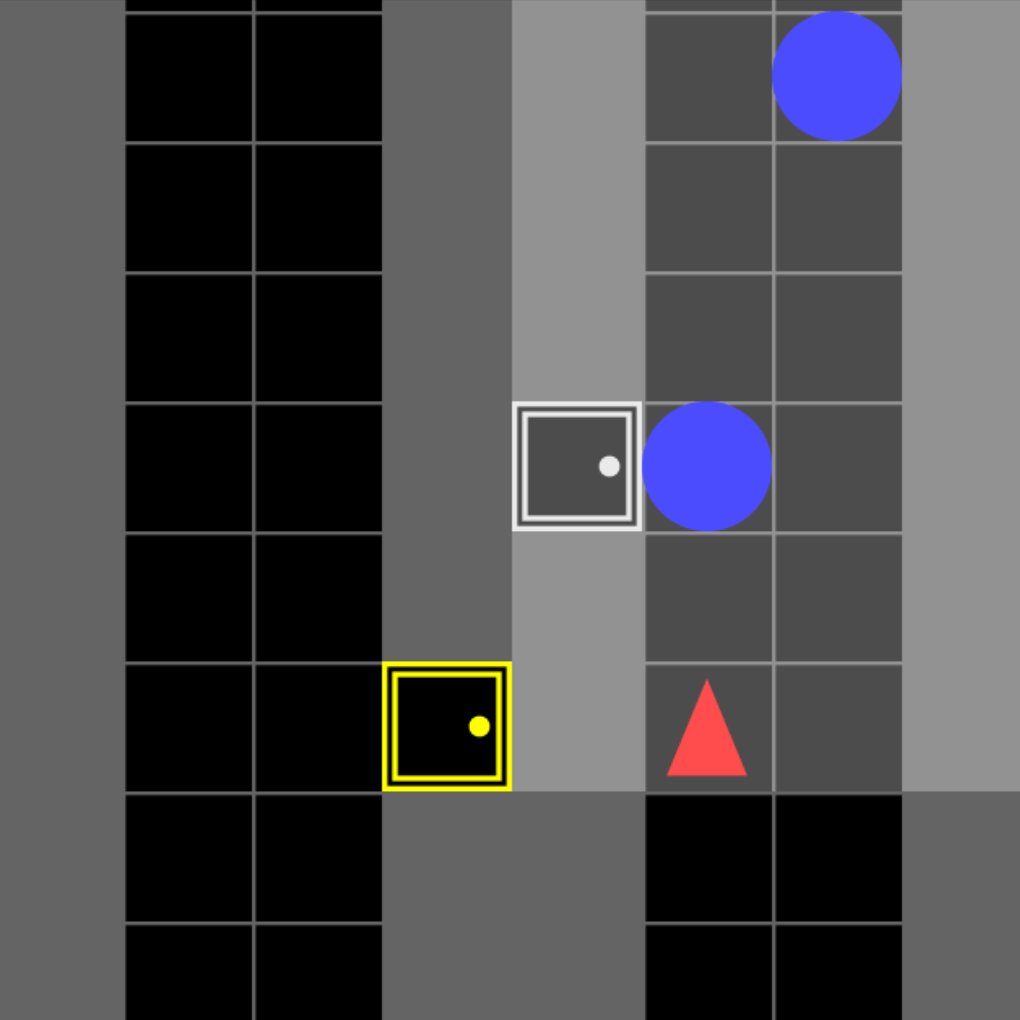}
    \captionof{figure}{\textbf{HeavyBall} and distractor doors in the \textbf{DualHallway} environment}
\end{center}

\noindent\textbf{Distractors}: For some \textbf{DualHallway} experiments, during test time we introduce new challenges for our behavior cloning agent it hasn't seen before to test \edit{CA-RRT's} generalization capabilities. First, we flip the hallway onto its side by rotating the whole thing by $90^{\circ}$. We also narrow the entire environment by one cell. This is the \textbf{SidewaysDualHallway} setup. Next, we introduce a setup with $5$ distractor doors at random locations in random colors along the middle wall. This is the \textbf{DualHallway + Distractor} setup. Finally, on top of the distractor doors we place $5$ \textit{immovable} \textbf{HeavyBall} items randomly into the environment. This is the \textbf{DualHallway + Distractors \& Obstacles} setup. This setup reduces the number of total states possible as a heavy ball can be in the path of exploration (see \edit{\hyperref[figure_ddb_example]{Figure 2}}).

\edit{H\edit{S}-RRT cannot be run in the \textbf{SidewaysDualHallway}, \textbf{DualHallway + Distractors}, or \textbf{DualHallway + Distractors \& Obstacles} setups as only one trajectory from one instantiation of the original \textbf{DualHallway} setup was collected. We introduce these additional setups mainly as a method for testing generalization abilities for CA-RRT {over novel maps}}.

\section{Results}
H\edit{S}-RRT was evaluated solely on the one instantiation of the environment where human data was collected. All other methods were evaluated on a large set of permutations of the environment: no two runs for CA-RRT or RRT were evaluated on the same instantiation. Even with just one trajectory from a human game player, CA-RRT improves the effectiveness with which we explore a game's state space. In all figures, colored areas represent $5$th to $95$th percentile, and the solid colored line represents the median.

% TODO: more explanations about what they're seeing, why things are missing
\subsection{DualHallway}

\begin{figure*}[h!]
    \centering
    \begin{subfigure}[b]{0.49\textwidth}
         \centering
         \includegraphics[width=\linewidth]{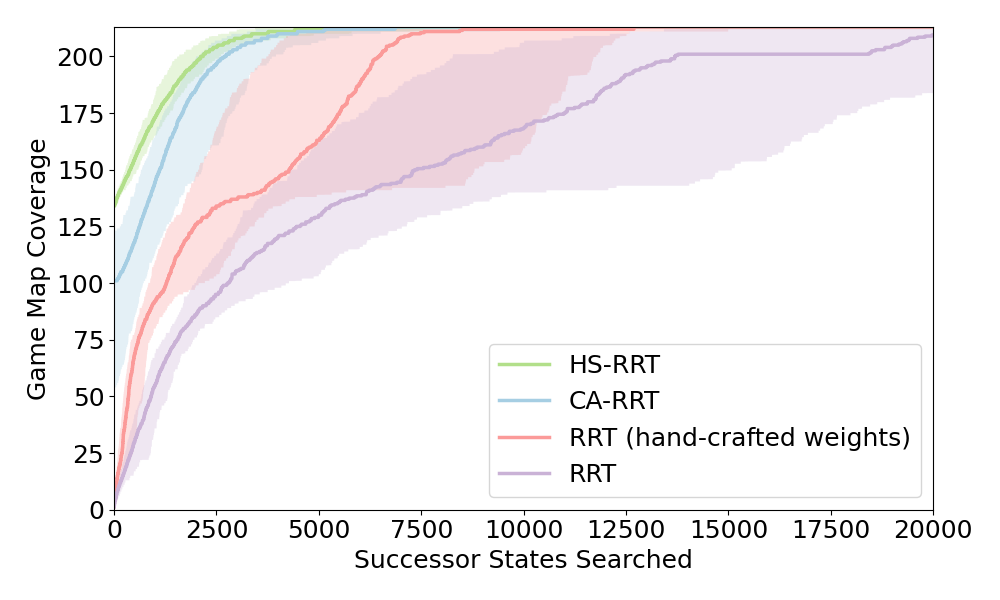} % 1
        %  \caption{(a)}
         \label{fig:hsrrtdual}
     \end{subfigure}
     \hfill
    \begin{subfigure}[b]{0.49\textwidth}
         \centering
         \includegraphics[width=\linewidth]{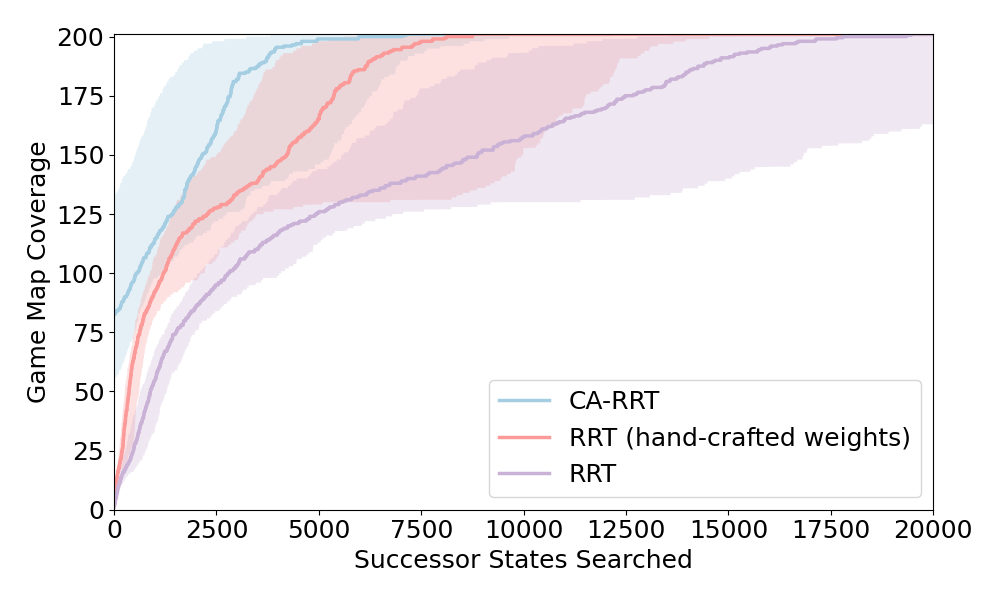} % 1
        %  \caption{(a)}
         \label{fig:sidewaysfigure}
     \end{subfigure}
    \caption{The plots show the total number of novel reachable states explored w.r.t. iterations of RRT expansions. (a) \textbf{DualHallway} - original setup.  {HS-RRT performs better in this task compared to all other algorithms because it is being tested on the same environment instance repeatedly. Our approach, CA-RRT, catches up with HS-RRT, but is slightly slower as it is using a trained policy and not a golden human-demonstration. 
    (b) \textbf{SidewaysDualHallway}. Our approach CA-RRT performs better in this task as it can generalize to novel environment instances with completely different orientations because it uses the human demonstrations to learn heuristics for search. HS-RRT cannot even be applied in such scenarios.}}
    \label{fig:my_label}
\end{figure*}

% \noindent\begin{minipage}{0.47\textwidth}
%     \begin{center}
%     \label{hsrrtdual}
%     \includegraphics[width=\textwidth]{imgs/hsrrt.png}
%     \captionof{figure}{\textbf{DualHallway} - original setup. The plot shows the total number of novel reachable states explored w.r.t. iterations of RRT expansions. We see that HS-RRT performs better in this task compared to all other algorithms because it is being tested on the same environment instance repeatedly.}
%     % The x-axis describes the number of iterations each RRT based method was run.}
%     \end{center}
% \end{minipage}
% \noindent \\

For the experiment on the \textbf{DualHallway} original setup (see \hyperref[fig:hsrrtdual]{Figure 3a}), H\edit{S}-RRT is seeded with {an expert human-demonstrated trajectory,} which in the original setup totalled $134$ out of the $213$ total states. CA-RRT, with no reliance on instantiations of the environment involving human-produced trajectories, was able to roll out and visit an average of $101$ states in the original setup to seed RRT search in the original setup, and on average saturated the game space a full $5831$ \edit{iterations} ($46\%$ fewer iterations) before weighted RRT.

% \noindent\begin{minipage}{0.47\textwidth}
%     \begin{center}
%     \label{sidewaysfigure}
%     \includegraphics[width=\textwidth]{imgs/sideways.png}
%     \captionof{figure}{\textbf{SidewaysDualHallway}:  The plot shows the total number of novel reachable states explored w.r.t. iterations of RRT expansions. We see that CA-RRT performs better in this task as it can generalize to novel environment instances with completely different orientations because it uses the human demonstrations to learn heuristics for search.}
%     \end{center}
% \end{minipage}
% \noindent \\

In the \textbf{SidewaysDualHallway} setup (see \hyperref[fig:sidewaysfigure]{Figure 3b}), CA-RRT average performance dips, where the average number of states visited during rollout was $83$, and fully saturated the game state space only $1617$ \edit{iterations} ($18\%$ fewer iterations) before weighted RRT did on average. However, only $94\%$ of weighted RRT experiments were able to saturate the game state space in $20,000$ \edit{iteration} compared to all $100\%$ of CA-RRT experiments. 
% \footnote{Note: HS-RRT is not run on \textbf{SidewaysDualHallway} because it requires collected trajectories and cannot generalize to unseen environments or unseen environment instantiations. \textbf{SidewaysDualHallway} was only made available during the evaluation phase}.

% \noindent\begin{minipage}{0.47\textwidth}
%     \begin{center}
%     \label{distractorfigure}
%     \includegraphics[width=\textwidth]{imgs/dd.png}
%     \captionof{figure}{\textbf{DualHallway + Distractors}}
%     \end{center}
% \end{minipage}

% \noindent\begin{minipage}{0.47\textwidth}
%     \begin{center}
%     \label{obstaclefigure}
%     \includegraphics[width=\textwidth]{imgs/ddb.png}
%     \captionof{figure}{\textbf{DualHallway + Distractors \& Obstacles}}
%     \end{center}
% \end{minipage}
% \noindent \\

\begin{figure*}[h!]
    \centering
    \begin{subfigure}[b]{0.49\textwidth}
         \centering
         \includegraphics[width=\linewidth]{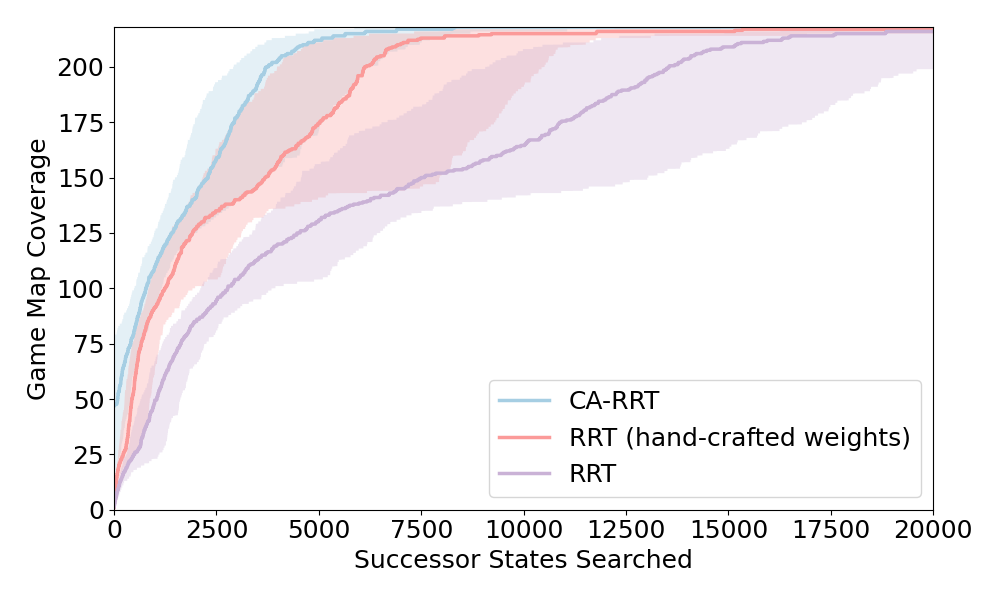} % 1
        %  \caption{(a)}
         \label{fig:distractorfigure}
     \end{subfigure}
     \hfill
    \begin{subfigure}[b]{0.49\textwidth}
         \centering
         \includegraphics[width=\linewidth]{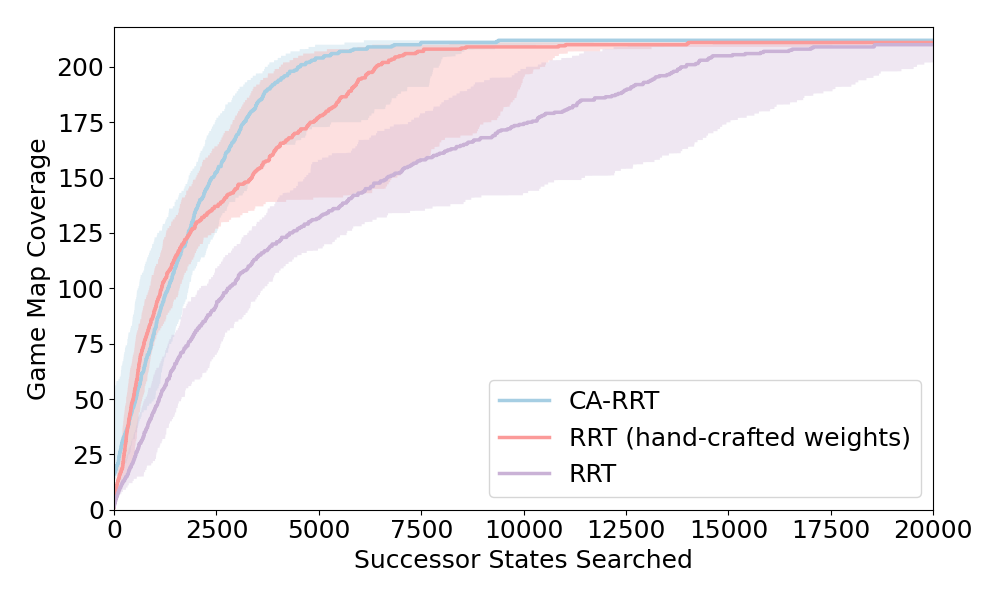} % 1
        %  \caption{(a)}
         \label{fig:obstaclefigure}
     \end{subfigure}
    \caption{The plot shows the total number of novel reachable states explored w.r.t. iterations of RRT expansions. (a) \textbf{DualHallway + Distractors}.  CA-RRT performs better in this task given its ability to generalize. {HS-RRT cannot be run in these environments as it is tethered to the environment instance where the human demonstrations were collected.}
    (b) \textbf{DualHallway + Distractors \& Obstacles}. CA-RRT performs better in this task than weighted RRT on average but advantage is reduced.}
    \label{fig:dualhallway-dd-b}
\end{figure*}

For both the \textbf{Distractors} (\hyperref[fig:distractorfigure]{Figure 4a}) and \textbf{Distractors \& Obstacles} (\hyperref[fig:obstaclefigure]{Figure 4b}), RRT with hand-crafted weights' performance curves changed minimally, but CA-RRT's performance takes a noticeable hit. When only adding distracting doors, we see CA-RRT saturates the game space in $7033$ fewer iterations than weighted RRT ($54\%$ fewer iterations) CA-RRT was to roll out nearly as far as in previous environments ($16$ states on average for \textbf{Distractors \& Obstacles}). On average, the CA-RRT algorithm saturates the state space more efficiently than the other two RRT methods, but its advantage throughout time has diminished significantly. Nevertheless, our experiments find in the \textbf{Distractors \& Obstacles} setup CA-RRT saturates the game state space $4636$ ($33\%$ fewer iterations) steps before weighted RRT does on average, suggesting weighted RRT struggled to expand to the final last few states more than CA-RRT. 
% \edit{Note: HS-RRT is not run on \textbf{DualHallway + Distractors} or \textbf{DualHallway + Distractors \& Obstacles} because it requires human collected trajectories from the test environment and cannot generalize to unseen environments or unseen environment instantiations; these setups were only made available during the evaluation phase}.

\subsection{CascadingLockDoor}

\begin{figure*}[h!]
    \centering
    \begin{subfigure}[b]{0.49\textwidth}
         \centering
         \includegraphics[width=\linewidth]{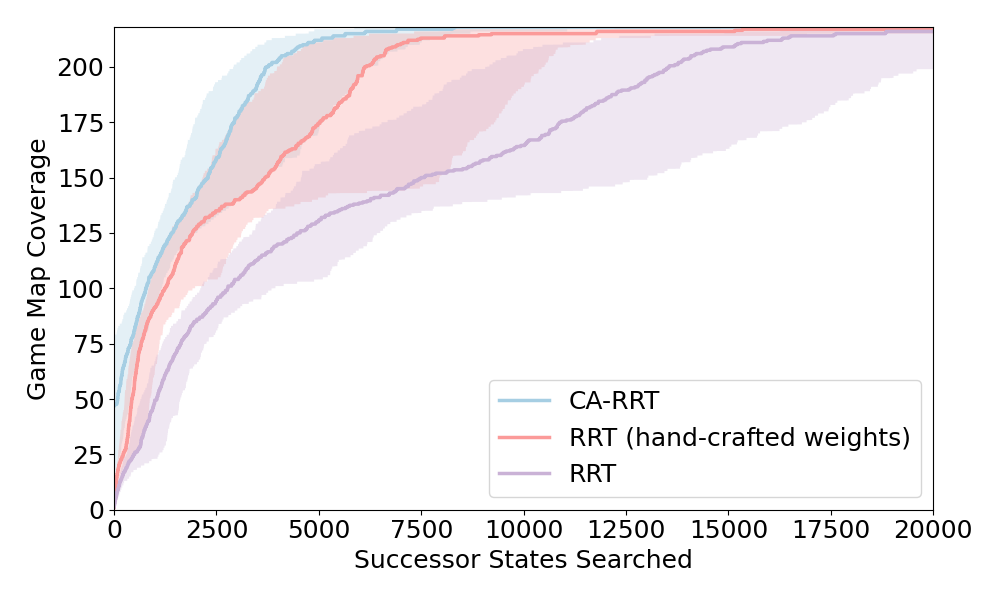} % 1
        %  \caption{(a)}
         \label{CLD}
     \end{subfigure}
     \hfill
    \begin{subfigure}[b]{0.49\textwidth}
         \centering
         \includegraphics[width=\linewidth]{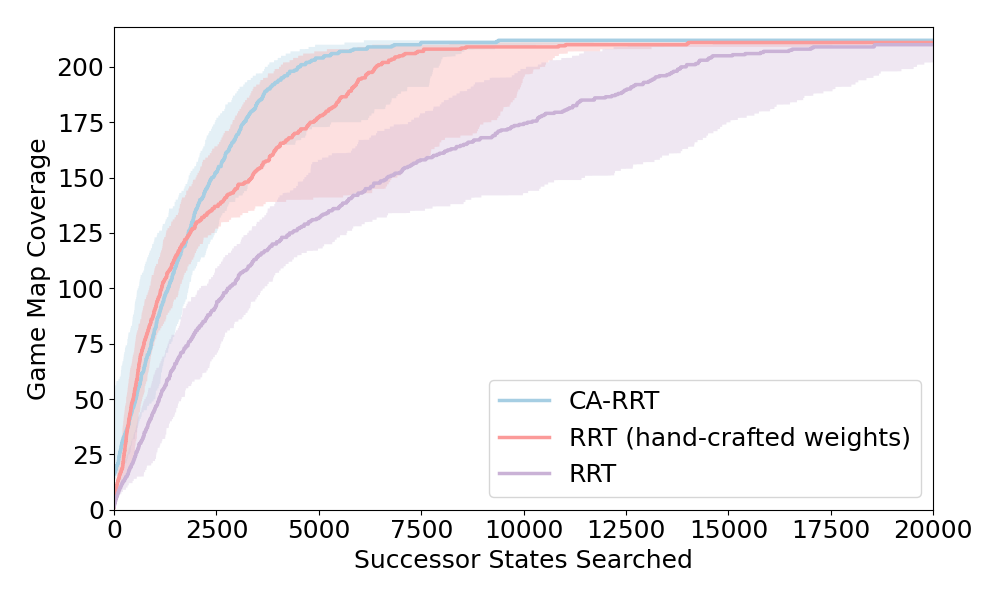} % 1
        %  \caption{(a)}
         \label{CLD2}
     \end{subfigure}
    \caption{The plot shows the total number of novel reachable states explored w.r.t. iterations of RRT expansions. (a) \textbf{CascadingLockDoor}. HS-RRT performs better in this task since human demonstrations initiailize RRT with access to both rooms immediately.
    (b) \textbf{CascadingLockDoor}. Same data as (a), but presenting only CA-RRT and weighted RRT for clarity.}
\end{figure*}

These graphs show the difficulty of reaching the second room (locked door is $37$th state\edit{: see y-value in \hyperref[CLD]{Figure 5a}} and \hyperref[CLD2]{Figure 5b}). \edit{The first plateau in \hyperref[CLD]{Figure 5a} is where the RRT methods got stuck trying to unlock the first room to go to the second room.}  Median statistics: H\edit{S}-RRT took the shortest amount of time, saturating the game state space in $1269$ \edit{iterations} ($90\%$ fewer iterations than weighted RRT). The CA-RRT algorithm reached the second room in $5933$ fewer \edit{iterations} ($47\%$ fewer iterations) than weighted RRT. RRT without weights was not able to reach the second room even after $20,000$ \edit{iterations} in the average case, remaining trapped.

\edit{CA-RRT, weighted RRT, and vanilla RRT all have drastically varying ranges} of performance, with CA-RRT's $5$th percentile performance worse than the median weighted RRT performance. Neither CA-RRT nor weighted RRT were able to solve the environment's last locked door in the $5$th-$95$th percentile range, although CA-RRT was able to solve it in $2\%$ of experiments, and weighted RRT did not solve it once. \edit{CA-RRT shows an improvement over weighted RRT, but still leaves significant more room to improve upon showing that these domains are challenging and require novel methods.}

\edit{\section{Discussion}}
\edit{
When evaluated on challenging custom gym-minigrid environments, it was found that  CA-RRT was able to \edit{reach} more \edit{meaningful states when} compared to the baseline weighted RRT~\cite{zhan2018taking} algorithm. {Moreover, our approach was more general purpose and applicable to novel maps when compared to the HS-RRT baseline~\cite{chang2019reveal}.} It does, however, remain possible to define more optimal hand-crafted weights, as we only chosen empirically. We showed that even with one human trajectory, the efficiency of exploration of a game's state space can be improved using CA-RRT.}

\edit{
However, there are several areas where CA-RRT falls short. First, real compute time: because our environments are very light, \edit{RRT iterations} are inexpensive. Further, since CA-RRT relies on a behavior cloning agent to produce action distributions at every step, using CA-RRT to explore may result in longer real compute time. In our experiments, when CA-RRT performed poorly, it took more than twice as long as RRT with hand-crafted weights running on a laptop. There are steps which can be taken that can reduce compute time (batch compute on GPU, store agent predictions, etc) which we have not yet explored. As the number of iterations correlates directly with the size of the tree computed, it is also possible that the cost of loading/saving all searched states becomes unsustainable.}

% \edit{\section{Future Work}}

% \edit{
% The minigrid environment is a very simple one: it was deterministic, discrete, limited interactable objects, and visual features were kept to a minimum. To fully realize the efficacy of our method, we would like to expand our tests to larger continuous environments such as SNES games even 3D games.}

% \edit{
% A reasonable area to expand to next is to adapt CA-RRT for efficient planning for robots. Robots may often have large degrees of freedom and large action spaces to search through. As demonstrated, CA-RRT was able to saturate a state space faster than weighted RRT given no specific goal condition. For example, if a goal condition was to reach a particular state in the second room of \textbf{CascadingLockDoor} but limited to no guidance could be provided, CA-RRT would be able provide a solution faster than either weighted RRT or vanilla RRT.
% }

% \edit{
% We would also like to incorporate more powerful learning from demonstration methods such as Inverse Reinforcement Learning to improve the action likelihood distribution weighting in CA-RRT.}

\section{Conclusion}
As games continue to grow in size, the work required to find bugs and flaws in the game state space grows, motivating for efficient automated game testing. Previous literature has explored the use of methods like weighted RRT for game testing\cite{zhan2018taking} {and the use of human demonstrations to seed the configuration states of RRT for exploration~\cite{chang2019reveal}}. As demonstrated, vanilla RRT can easily produce a trapped agent if not assisted in some form {and plain seeding with global states of a game reduces the applicability of the approach to novel maps}. We find CA-RRT to be more useful in covering game state space for testing compared to the baseline RRT method proposed in previous literature~\cite{zhan2018taking} and {the human-seeded RRT approach~\cite{chang2019reveal}}. \edit{While the focus of this paper is efficient wide state space exploration and saturation, we hope that an adapted CA-RRT method, like other similar human-guided RRT methods~\cite{griner2012human}~\cite{caves2010human} will be able to also help robots find goal conditions more efficiently}.

\bibliographystyle{plain}
\bibliography{references}
% \flushends
% \newpage
%%%%%%%%%%%%%%%%%%%%%%%%%%%%%%%%%%%%%%%%%%%%%%%%%%%%%%%%%%%%

% \section{Appendix}
% \label{Appendix}

% \subsection{DualHallway Figures}
% \begin{center}
%     \includegraphics[width=\textwidth]{imgs/carrt_dh_all.png}
%     \captionof{figure}{All methods compared}
% \end{center}
% \begin{center}
%     \includegraphics[width=\textwidth]{imgs/carrt_dh_nd.png}
%     \captionof{figure}{Non-deterministic methods methods compared}
% \end{center}

% \subsection{CascadingLockDoor Figures}
% % \begin{center}
% %     \includegraphics[width=\textwidth]{imgs/carrt_cl.png}
% %     \captionof{figure}{All methods compared}
% % \end{center}
% \begin{center}
%     \includegraphics[width=\textwidth]{imgs/carrt_cl_nd_only.png}
%     \captionof{figure}{Non-deterministic methods methods compared}
% \end{center}
% % Optional

\end{document}